\documentclass{article}
\pdfoutput=1
\usepackage{microtype}
\usepackage{graphicx}
\usepackage{subfigure}
\usepackage{booktabs} % for professional tables

\usepackage{hyperref}

\usepackage{multirow}
\usepackage{booktabs}       % professional-quality tables
\usepackage{amsmath}
\usepackage{amsfonts}
\usepackage{xcolor}

\newcommand{\TT}[1]{\texttt{#1}}
\newcommand{\BF}[1]{\textbf{#1}}
\newcommand{\IT}[1]{\textit{#1}}

\newcommand{\UL}[1]{\underline{#1}}

\newcommand{\M}[1]{\mathbf{#1}}

\newcommand{\floor}[1]{\left\lfloor #1 \right\rfloor}

\usepackage[accepted]{icml2019}

\icmltitlerunning{Improving Neural Network Quantization using Outlier Channel Splitting}

\begin{document}

\twocolumn[
\icmltitle{Improving Neural Network Quantization without Retraining using \\ Outlier Channel Splitting}



\icmlsetsymbol{equal}{*}

\begin{icmlauthorlist}
\icmlauthor{Ritchie Zhao}{c}
\icmlauthor{Yuwei Hu}{c}
\icmlauthor{Jordan Dotzel}{c}
\icmlauthor{Christopher De Sa}{c}
\icmlauthor{Zhiru Zhang}{c}
\end{icmlauthorlist}

\icmlaffiliation{c}{Cornell University, Ithaca, New York 14850, USA}

\icmlcorrespondingauthor{Ritchie Zhao}{rz252@cornell.edu}
%\icmlcorrespondingauthor{Eee Pppp}{ep@eden.co.uk}

\icmlkeywords{Machine Learning}

\vskip 0.3in
]



\printAffiliationsAndNotice{}  % leave blank if no need to mention equal contribution

\begin{abstract}
Quantization can improve the execution latency and energy efficiency of neural networks on both commodity GPUs and specialized accelerators.
The majority of existing literature focuses on training quantized DNNs, while this work examines the less-studied topic of quantizing a floating-point model without (re)training.
DNN weights and activations follow a bell-shaped distribution post-training, while practical hardware uses a linear quantization grid.
This leads to challenges in dealing with \IT{outliers} in the distribution.
Prior work has addressed this by clipping the outliers or using specialized hardware.
In this work, we propose outlier channel splitting (OCS), which duplicates channels containing outliers, then halves the channel values. The network remains functionally identical, but affected outliers are moved toward the center of the distribution.
OCS requires no additional training and works on commodity hardware.
Experimental evaluation on ImageNet classification and language modeling shows that OCS can outperform state-of-the-art clipping techniques with only minor overhead.
\end{abstract}

\section{Introduction}
\label{sec:intro}

Over the past few years, deep neural networks (DNNs) have become the state-of-the-art approach for many large-scale computer vision and sequence modeling problems. Deep convolutional networks dominate the leaderboards for popular image classification and object detection datasets such as ImageNet~\citep{deng2009imagenet} and Microsoft COCO~\citep{lin2014coco}.
However, the significant compute and memory requirements of running DNNs impedes the adoption of neural nets in application domains such as edge computing or latency-critical services~\citep{xu2018scaling}.
One approach to reducing the costs of DNN execution is to quantize the floating-point weights and activations into low-precision fixed-point numbers.
This reduces the model size as well as the complexity of multiply-accumulate (MAC) operations in hardware, enabling better throughput and energy efficiency.
%This has been shown to improve DNN throughput and energy efficiency on both GPUs and specialized accelerators.
DNN quantization is an active area of research~\citep{wu2018training,jacob2018quantization,choi2018pact,banner2018aciq} and sees deployment in commercial systems such as Google's TPU~\citep{google2017tpu}, NVIDIA's TensorRT~\citep{tensorrt2017slides}, and Microsoft's Brainwave~\citep{microsoft2018brainwave}.

The majority of literature on DNN quantization involves training --- either from scratch
~\citep{courbariaux2015binaryconnect,wu2018training,jacob2018quantization}
or retraining/fine-tuning from a floating-point model.
~\citep{han2016deep,zhou2017incremental}.
Although such techniques are valuable, there are important real-world scenarios in which (re)training is not applicable.
Consider an ML service provider (e.g. Amazon, Microsoft, Google) which wants to run a black-box  floating-point client model in low-precision. The service provider does not have the training data, and the client may not be able to train for quantization because: (1) it lacks the expertise or manpower; (2) it is using an off-the-shelf or legacy model for which training data is not available.
The importance of \IT{post-training} quantization can be seen from NVIDIA's TensorRT, a product specifically designed to perform 8-bit integer quantization without (re)training. This paper focuses on post-training DNN quantization.

\begin{figure*}[t]
\vskip 0.2in
  \begin{center}
    %\hfill
    %\begin{minipage}{0.33\textwidth}
    %  \centerline{\includegraphics[width=\columnwidth, trim=20px 10px 30px 20px]{linear6.png}}
    %\end{minipage}
    %\hfill
    %\begin{minipage}{0.33\textwidth}
    %  \centerline{\includegraphics[width=\columnwidth, trim=20px 10px 30px 20px]{clip6.png}}
    %\end{minipage}
    %\hfill
    %\begin{minipage}{0.33\textwidth}
    % \centerline{\includegraphics[width=\columnwidth, trim=20px 10px 30px 20px]{ocs6.png}}
    %\end{minipage}
    %\hfill
    \hfill
    \begin{minipage}{0.99\textwidth}
      \centering
      \includegraphics[width=\columnwidth]{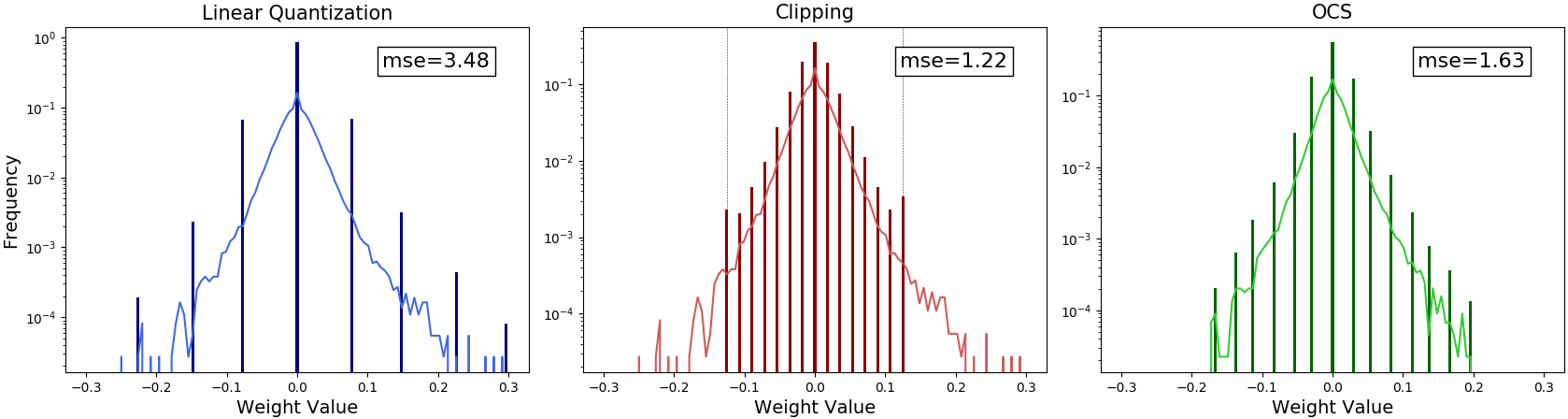}
    \end{minipage}
    \hfill
    \vskip 0.05in
    \hfill
    \begin{minipage}{0.32\textwidth}
      \centering \hskip 0.3in (a)
    \end{minipage}
    \hfill
    \begin{minipage}{0.32\textwidth}
      \centering \hskip 0.3in (b)
    \end{minipage}
    \hfill
    \begin{minipage}{0.32\textwidth}
      \centering \hskip 0.3in (c)
    \end{minipage}
    \hfill
    \vspace{-0.05in}
    \caption{\BF{Weight histograms for linear, clipping, and OCS quantization techniques}.
    The floating-point weight histogram is in the light color while the quantized weight histogram is in dark color.
    Both clipping and OCS reduces mean squared quantization error (MSE) by reducing the dynamic range of the distribution. Clipping reduces the overall MSE but greatly distorts the outliers. OCS avoids this distortion by splitting the outliers instead, moving them towards the center of the distribution at the cost of some model size overhead.
    }
    \label{fig:distributions}
  \end{center}
\vskip -0.2in
\end{figure*}

DNN weights and activations follow a bell-shaped distribution after training. However, commodity hardware uses a linear number representation with evenly-spaced grid points.
The na\"{\i}ve approach is to linearly map the entire range of the distribution to the range of the quantization grid (Figure~\ref{fig:distributions}(a)). Here the grid points extend to the maximum value in the distribution~\citep{hubara2017quantized}.
Clearly, this method over-provisions grid points for the rarely-occurring \IT{outliers}.
A better approach is to make the grid narrower than the distribution --- this is known as \IT{clipping}, as it is equivalent to thresholding the outliers before applying linear quantization (Figure~\ref{fig:distributions}(b)).
Empirically, clipping can improve the accuracy of quantized DNNs, and many techniques exist to choose the optimal clip threshold~\citep{sung2015resiliency,zhuang2018towards,tensorrt2017slides}.
Unfortunately, clipping can only reduce overall quantization error by increasing the distortion on the outliers --- it is constrained by this tradeoff.

Another approach to handling outliers is to quantize them separately from the central values. Such outlier-aware quantization~\citep{park2018outlier,park2018value} is highly effective, but involves the use of dedicated non-commodity hardware.

In this paper, we propose \IT{outlier channel splitting} (OCS). OCS identifies a small number of channels containing outliers, duplicates them, then halves the values in those channels. This creates a functionally identical network, but moves the affected outliers towards the center of the distribution (Figure~\ref{fig:distributions} (c)).
OCS takes inspiration from Net2Net~\citep{chen2015net2net}; it does not require retraining and can be used on commodity CPUs and GPUs.
OCS introduces a new tradeoff: it reduces quantization error at the expense of making the neural network larger.
Experimental evaluation shows that for practical CNN and RNN models, OCS can significantly improve post-training quantization accuracy over state-of-the-art clipping methods with just a few percent overhead.

To present a comprehensive study of post-training quantization, we also evaluate different techniques for optimizing the clip threshold on both weights and activations. To our best knowledge, we are the first to perform a detailed literature comparison.
Code for both OCS and clipping is available in open source~\footnote{https://github.com/cornell-zhang/dnn-quant-ocs}.
Our specific contributions are as follows:
\begin{enumerate}
\itemsep0em 
    \item We propose outlier channel splitting, a technique to improve DNN model quantization that does not require retraining and works with commodity hardware.
    \item We present a comprehensive evaluation of post-training clipping techniques found in literature. To our best knowledge this is the first such study.
    \item We demonstrate that OCS can outperform state-of-the-art clipping techniques on weight quantization, while incurring negligible overheads.
\end{enumerate}

\section{Related Work}
\label{sec:related}

\subsection{Post-Training Quantization}
Clipping is the state-of-the-art for DNN quantization without training. Clipping can be applied to both weights and activations --- for the latter the activation distributions are sampled from a small number of inputs.
\citep{sung2015resiliency} and \citep{shin2016fixed} examined post-training quantization for CNNs and RNNs, respectively. They adopt a clip threshold that minimizes the L2-norm of the quantization error.
ACIQ~\citep{banner2018aciq} fits a Gaussian and Laplacian to the sampled distribution, then uses the better-fitting curve to analytically compute the optimal clip threshold.
In a similar vein, SAWB~\citep{choi2018qnn} linearly extrapolates the clip threshold using statistics from fitting six different distributions.
\citep{mckinstry2018clip} clips using a percentile of the sampled values, the exact percentile depends on the quantization bitwidth.
Different from the others, \citep{settle2018quantizing} tunes the bitwidth and floating-point format, achieving 32-bit accuracy performance with only 8-6 bits.
% post-training, bitwidth tuning, FP format optimization

%learning clip threshold using EMA~\citep{jacob2018quantization}
%learning clip threshold using backprop~\citep{choi2018pact}
NVIDIA's TensorRT~\citep{tensorrt2017slides} is a commercial library that quantizes floating-point models to 8-bit for GPU inference.
Clipping is used for the activations to control the effect of outliers.
TensorRT profiles the activation distributions using a small number (1000s) of user-provided training samples, then computes a clipping threshold by minimizing the KL divergence between the original and quantized distributions.
%Both TensorRT and OCS address quantization without retraining. However, we target lower precisions than 8-bit which require additional techniques beyond clipping.

OCS is different from these works as it leverages model expansion to improve quantization.

\subsection{Outlier-Aware Quantization}
Park et al. propose outlier-aware quantization~\citep{park2018value,park2018outlier}, which uses a low-precision grid for the center values and a high-precision grid for the outliers. Placing ~3\% of values on the high-precision grid enabled post-training quantization of many popular CNN models to 4-bit without accuracy loss.
This technique requires a specialized outlier-aware DNN accelerator; our approach is very different as it is designed to be applicable on commodity hardware.

\subsection{Net2Net}
OCS is inspired by Net2Net~\citep{chen2015net2net}, which presents transformations to make a neural network wider or deeper while preserving functional equivalence.
The goal of Net2Net was to speed up training by expanding a smaller DNN model into a larger one; the larger model inherits knowledge from the smaller and does not need to be trained from scratch.
In this work we apply the \TT{Net2WiderNet} transform to reduce outliers and improve quantization.

\subsection{Cell Division}
Cell division~\citep{park2019celldiv} examines the same idea as OCS, and was published concurrently with our work. Though conceptually very similar, there are some technical differences between their work and ours: (1) they apply OCS on weights only while we examine both weights \IT{and activations}; (2) they first tune the fixed-point bitwidths per layer with 50K training images while we use no data for weight OCS; (3) they do not compare against clipping as a baseline; (4) they evaluate on MNIST, CIFAR-10, and AlexNet while we evaluate on more modern (i.e. post-ResNet) ImageNet CNNs and language models.
\begin{figure*}[t]
\vskip 0.1in
  \begin{center}
    \hfill
    \begin{minipage}{0.5\textwidth}
      \centering
      \includegraphics[width=\columnwidth]{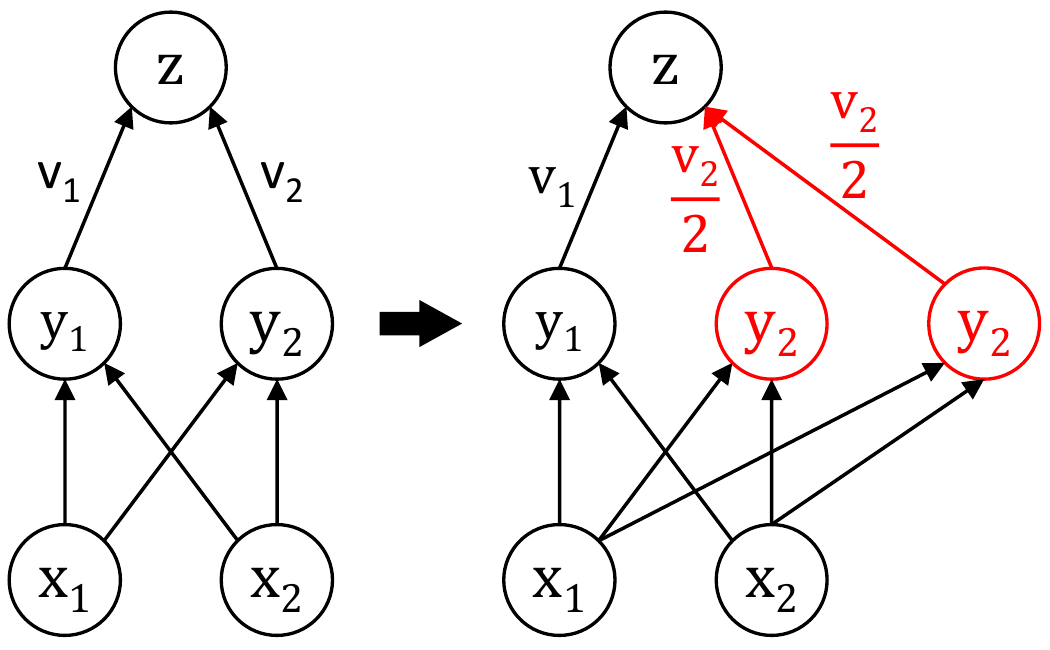}
    \end{minipage}
    \hfill
    \begin{minipage}{0.35\textwidth}
      \centering
      \includegraphics[width=\columnwidth]{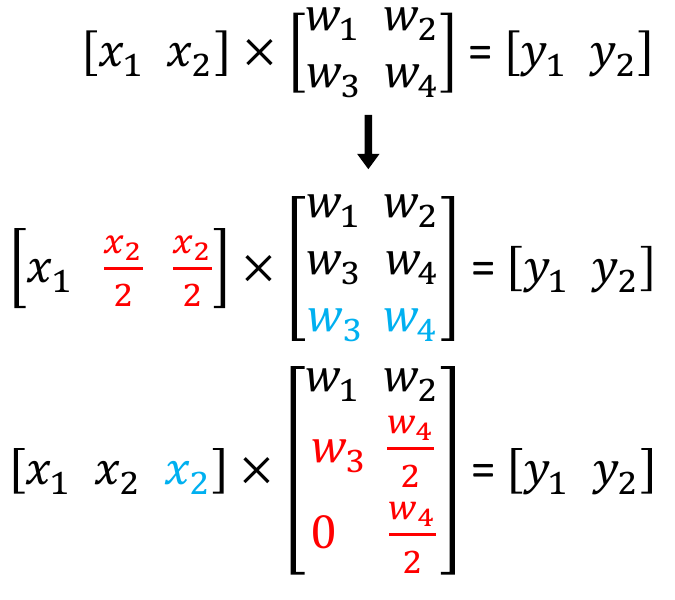}
    \end{minipage}
    \hspace{0.5in}
    \hfill
    %\vspace{0.05in}
    \hfill
    \begin{minipage}{0.55\textwidth}
      \centering
      (a)
    \end{minipage}
    \hfill
    \begin{minipage}{0.37\textwidth}
      \centering
      \hspace{0.15\textwidth}(b)
    \end{minipage}
    \hfill
    \vskip -0.1in
    \caption{\BF{OCS network transformation --} after duplicating a neuron, we can divide either the neuron's output value or its outgoing weights in half to preserve functional equivalence. (a) figure taken from Net2Net~\cite{chen2015net2net} where the weight $v_2$ is halved by duplicating $y_2$;
    (b) an example with multiple inputs and multiple outputs, showing how $x_2$ or $w_3$ can be halved while maintaining the same outputs. Here, an entire row must be added to the weight matrix to split a single value.
    }
    \label{fig:net2widernet}
  \end{center}
\vskip -0.1in
\end{figure*}

\section{Outlier Channel Splitting}
\label{sec:ocs}

\subsection{Linear Quantization}
The simplest form of linear quantization maps the inputs to a set of discrete, evenly-spaced grid points which span the entire dynamic range of the inputs. The maximum quantization error for any single value is one-half of the increment. For symmetric $k$-bit quantization, we have $2^{k}-1$ grid points:
\begin{equation}
    \text{\BF{LinearQuant}}(\M{x}) = \text{round}\bigg(\frac{\M{x}\,(2^{k-1}-1)}{\max(|\M{x}|)}\bigg)\frac{\max(|\M{x}|)}{2^{k-1}-1}
    \label{eqn:linearquant}
\end{equation}
Because each value is scaled by $\max(\M{x})$, \BF{LinearQuant} is sensitive to the largest inputs, i.e. the outliers.
Many existing works first clip the range of $\M{x}$ prior to linear quantization; a survey of such clipping techniques can be found in Section~\ref{sec:clipping}.

\subsection{Improving Quantization with \TT{Net2WiderNet}}
\label{sec:ocs-basic}
The core idea of OCS is to reduce the magnitude of outlier weights and/or activations in a DNN layer by duplicating a neuron, then either (1) halving its output; (2) halving the outgoing weight connections. This leaves the layer functionally equivalent but makes the weight/activation distribution narrower and thus more suitable for linear quantization. Such layer transformations were originally proposed as \TT{Net2WiderNet} in Net2Net~\citep{chen2015net2net}; we leverage them to improve quantization.

More formally, consider a linear layer in a DNN which takes as input the $m$-channel activation vector $\M{x} = \{\M{x}_i\}_{i=0}^m$, where each $\M{x}_i$ can be a single value (FC layer) or a 2D feature map (conv layer). Let $\M{y} = \{\M{y}_j\}_{j=0}^n$ be the $n$-channel output.
We can define a linear layer as follows:
\begin{equation}
    \M{y}_j = \sum_{i=1}^{m}\M{x}_i * \M{W}_{ij} \label{eqn:conv}
\end{equation}
where $\M{W}_{ij}$ represents the weight(s) connecting $\M{x}_i$ and $\M{y}_j$ and $*$ represents multiplication or 2D convolution over a single channel.
Without loss of generality, consider using OCS to split the last channel $\M{x}_m$.
This equates to rewriting Equation~\ref{eqn:conv} as follows:
\begin{align}
    \M{y}_j &= \sum_{i=1}^{m-1}\M{x}_i * \M{W}_{ij} + (\M{x}_m * \frac{\M{W}_{mj}}{2}) + (\M{x}_m * \frac{\M{W}_{mj}}{2}) \label{eqn:wt-ocs} \\
    \M{y}_j &= \sum_{i=1}^{m-1}\M{x}_i * \M{W}_{ij} + (\frac{\M{x}_m}{2} * \M{W}_{mj}) + (\frac{\M{x}_m}{2} * \M{W}_{mj}) \label{eqn:act-ocs}
\end{align}
In both cases, we split channel $m$ into 2 channels. To preserve equivalence, we can halve the weights (Equation~\ref{eqn:wt-ocs}) or halve the input activations (Equation~\ref{eqn:act-ocs}).
Figure~\ref{fig:net2widernet}(a) taken from the Net2Net paper illustrates weight OCS visually: 
by duplicating $y_2$ we can cut its outgoing weight $v_2$ in half.

OCS is an alternative to clipping for reducing the dynamic range of DNN values without retraining. Compared to clipping, OCS preserves the outliers but incurs additional network overhead. The outlier values are the largest values in a layer and contribute the most to the outputs. We expect OCS to outperform clipping in neural network accuracy --- the question is whether it can do so with low overhead.

Figure~\ref{fig:net2widernet}(b) shows some additional caveats of OCS in a layer with 2 inputs and 2 outputs. The top equation describes the original layer; the next two equations illustrate OCS to split the activations and the weights, respectively.
One caveat is that to split any weight value, an entire row must be added to the weight matrix. For a conv layer, OCS requires duplicating an entire 2D activation channel and all 2D weight filters connected to that channel.
A second caveat is that not all values need to be split. At the bottom of Figure~\ref{fig:net2widernet}(b), $w_4$ is split in half while $w_3$ is not split.
%This can be leveraged to minimize the amount of edits made to the layer's values.

\subsection{Quantization-Aware Splitting}
\label{sec:q-aware}
In this section, we show that duplicating a value and dividing it by two (i.e. \TT{Net2WiderNet}) increases the total quantization error. We then propose an alternative split ratio which preserves quantization error.
Without loss of generality, consider the deterministic rounding function $Q(x) = \floor{x + \frac{1}{2}}$, which maps each real number to its closest integer, rounding halves toward $-\infty$.
The maximum quantization error introduced by $Q(x)$ is $0.5$.
Next define OCS as a function $f(w): \mathbb{R} \rightarrow \mathbb{R}^2$ which maps a single value to two values. The na\"{\i}ve split used in \TT{Net2WiderNet} is:
\begin{equation}
    \text{OCS}_{\text{naive}}(w) = \begin{pmatrix}
    w/2 \\
    w/2\end{pmatrix}
\end{equation}
It is clear that $Q(w) \neq Q(\frac{w}{2}) + Q(\frac{w}{2})$, i.e. na\"{\i}ve OCS does not preserve the quantized value. The maximum total quantization error is doubled as both halves may be rounded in the same direction (e.g., $w=3$ and each half is $1.5$).

To address this, we propose the following \IT{quantization-aware} (QA) splitting function:
\begin{equation}
    \text{OCS}_{\text{QA}}(w) = \begin{pmatrix}
    (w-0.5)/2 \\
    (w+0.5)/2\end{pmatrix}
\end{equation}
Intuitively, this forces $Q(x)$ to round in different directions when $w$ is close to the midpoint between grid points. More formally, we can prove the following:
\begin{equation}
\begin{aligned}
    &\,Q(\frac{w-0.5}{2}) + Q(\frac{w+0.5}{2}) \\
    = &\floor{\frac{w-0.5}{2}+\frac{1}{2}} + \floor{\frac{w+0.5}{2}+\frac{1}{2}} \\
    = &\floor{\frac{w+0.5}{2}} + \floor{\frac{w+0.5}{2} + \frac{1}{2}} \\
    = &\floor{w + 0.5}
\end{aligned}
\end{equation}
The last line is simply $Q(w)$, showing that QA OCS preserves the original quantization result.
To derive the last line, we apply Hermite's Identity~\citep{hermites} with $n=2$:
\begin{equation}
    \sum_{k=0}^{n-1}\floor{x + \frac{k}{n}} = \floor{nx}
\end{equation}

We can further show that QA splitting is optimal, i.e. there exists no way to split $w$ which results in lower quantization error. This proof is omitted due to length.

\subsection{Channel Selection}
As stated earlier, OCS cannot target individual weights or activations and must duplicate entire channels.
OCS performs splits one at a time, and always splits the channel containing the largest absolute value in the layer. By prioritizing channels containing the largest values, OCS seeks to minimize distortion caused by any subsequent clipping.

We use a simple method to determine how many splits to perform in each layer (i.e. how many extra channels are created). For a layer containing $C$ channels, OCS splits $\text{ceil}(r*C)$ channels, where $r$ is the \IT{expansion ratio}, a hyperparameter that determines approximately the level of tolerable overhead in the network.
This method allocates extra channels without considering each layer's weight or activation distributions. We also tried a more intelligent approach which formulates extra channel allocation as a knapsack problem. The reward function is the percentage reduction in the dynamic range of the distribution, and the cost is the increase in memory size. We optimize the number of extra channels for all layers simultaneously subject to a constraint on the memory overhead. Unfortunately, the knapsack approach is experimentally not better than the simple method described above, and for space reasons we do not show results with knapsack.

Channel selection on DNN weights is straightforward to implement as the weights are known and fixed post-training.
For the activations, we take an approach similar to TensorRT~\citep{tensorrt2017slides}: we use a small number of training images to sample the activations in each layer. The sampled distributions in each layer are then used for OCS.
%We find the $\text{ceil}(r*C)$ channels in which outliers occur with the highest frequency, (where outlier in this context is a value greater than the 99th-percentile of activations).
%We also tested metrics such as prioritizing channels with the largest max value, mean value, or variance.
%Preliminary experiments showed that regardless of channel selection method, statically choosing which channels to split was not effective. See additional discussion in Section~\ref{sec:exp-acts}.

\subsection{Implementation on Commodity Hardware}
\label{sec:ocs-impl}
A key strength of OCS is simplicity, allowing it to be used in practical scenarios with either commodity hardware or emerging deep learning accelerators.
Figure~\ref{fig:net2widernet}(b) shows the network modifications needed to implement OCS --- we need to duplicate and possibly scale certain channels in the weights and activations.
The weight modifications can be done off-line prior to serving the model.
For the activations, a custom layer can be inserted which simply copies and scales the appropriate channels.
%Note that this is more efficient that Net2Net, which relies on expanding the width of each layer.
%OCS is implemented on a commodity GPU as part of the experiments, and we believe OCS will be easy to implement on specialized DNN accelerators.
\section{Clipping}
\label{sec:clipping}
Clipping represents the state-of-the-art in post-training quantization. This section gives a brief overview of different methods for optimizing the clip threshold in literature; we present an evaluation of these methods in Section~\ref{sec:experiment}.

\subsection{Mean Squared Error}
This method chooses a clip threshold which minimizes the mean squared error (MSE) or L2-norm between the floating-point and quantized values~\citep{sung2015resiliency,shin2016fixed}.
It first constructs a histogram of the floating-point values. Let $x_i$ and $h(x_i)$ be the bin values and frequencies, and $i = 1 \dots n$ denote the $n$ bins. The MSE is defined as:
\begin{equation}
    \text{MSE} = \frac{1}{n}\sum_{i=1}^n h(x_i) * (x_i - Q(x_i))^2
\end{equation}
where $Q(x)$ is the quantization function.
In our experiments, we generate a large number of candidate clip thresholds evenly spaced between $0$ and the max absolute value, and choose the one with minimal MSE.

\subsection{ACIQ}
Proposed by~\citep{banner2018aciq}, ACIQ first determines whether distribution is closer to a Gaussian or a Laplacian.
Using statistics from the appropriate distribution, it uses an (approximate) closed-form solution for the clip threshold which minimizes MSE.
Compared to the MSE method above, ACIQ avoids sweeping candidate thresholds and is much faster --- this allows the clip threshold to be adjusted between input batches for activation quantization.

We used open-source code from the authors~\footnote{https://github.com/submission2019/AnalyticalScaleForInteger Quantization}.
Banner \IT{et al}. assumed that an $m$-bit fixed-point format contains $2^m$ grid points; this representation lacks a grid point at zero for signed values. We use $2^m-1$ grid points instead (i.e. sign-magnitude) as it is the default in our framework, and slightly adjusted the formulas from the paper to suit.

\subsection{KL Divergence}
This method chooses a clip threshold which (approximately) minimizes the KL divergence between the floating-point and quantized. Similar to the MMSE method, it works on the histogram of values and selects the optimal clip threshold from a set of candidates.
The method was first proposed in a set of slides on NVIDIA's TensorRT~\citep{tensorrt2017slides}, which unfortunately does not contain enough technical detail for replication.
Instead, we adapted an open-source implementation from Apache MXNet~\citep{chen2015mxnet}.

In general, floating-point and quantized distributions do not have the same support and the KL divergence is thus undefined. To get around this, the MXNet implementation smooths the quantized histogram slightly by moving some of the probability mass into zero-frequency bins.
\begin{table}[t]
  \centering
  \vskip -0.05in
  \begin{small}
      \caption{\BF{Quantization-aware (QA) splitting in OCS --} each table entry is formatted as (\BF{QA / non-QA}) where the non-QA split is simply dividing by two. The model is ResNet-20 for CIFAR-10.}
    \vskip 0.05in
    \begin{tabular}{c|c|c|c|c}
    \toprule
    \multicolumn{1}{c|}{\multirow{2}[0]{*}{\shortstack{\BF{Wt.}\\\BF{Bits}}}} & \multicolumn{4}{c}{\BF{OCS Expand Ratio}} \\
          & \multicolumn{1}{c}{\BF{0.01}} & \multicolumn{1}{c}{\BF{0.05}} & \multicolumn{1}{c}{\BF{0.1}} & \multicolumn{1}{c}{\BF{0.2}} \\
    \midrule
    6     & \BF{92.0} / 91.9  & 91.9 / \BF{92.0}  & \BF{92.1} / 91.9  & 92.0 / 92.0  \\
    5     & \BF{91.6} / 91.4  & \BF{91.7} / 91.6  & \BF{92.0} / 91.8  & 91.7 / \BF{91.8}  \\
    4     & 88.0 / \BF{88.3}  & 88.2 / \BF{88.3}  & \BF{88.7} / 86.8  & \BF{89.1} / 86.8  \\
    3     & \BF{49.9} / 44.5  & \BF{58.3} / 44.8  & \BF{62.7} / 44.6  & \BF{76.5} / 52.8  \\
    \bottomrule
    \end{tabular}%
    \label{tab:q-aware}%
  \end{small}
  \vskip -0.2in
\end{table}%

\section{Experimental Evaluation on CNNs}
\label{sec:experiment}

\newcommand{\HL}[1]{\textcolor[rgb]{ .29,  .525,  .91}{\UL{\BF{#1}}}}

% Table generated by Excel2LaTeX from sheet 'Paper'
\begin{table*}[t]
  \centering
  \vskip -0.1in
  \begin{small}
  \caption{\BF{ImageNet Top-1 validation accuracy with weight quantization --} the float accuracy is displayed under the model name.
  Results include different \BF{Clip} methods, \BF{OCS} with different expand ratios, and \BF{OCS + the Best Clip method} at each bitwidth.
  For clipping, the best result is bolded and copied to the \BF{Clip - Best} column.
  For OCS, the smallest expand ratio that outperforms \UL{all} clipping methods is bolded, and the smallest expand ratio that achieves +1\% accuracy over clipping is highlighted in blue. 
  Best viewed in color.
  }
  \vspace{3px}
    \begin{tabular}{c|c||cccc|c||ccc||ccc}
    \toprule
    \multicolumn{1}{c|}{\multirow{2}[0]{*}{\BF{Network}}} & \BF{Wt} & \multicolumn{4}{c|}{\BF{Clip}} & \BF{Clip} & \multicolumn{3}{c||}{\BF{OCS}} & \multicolumn{3}{c}{\BF{OCS + Best Clip}} \\
    & \BF{Bits} & \multicolumn{1}{c}{None} & \multicolumn{1}{c}{MSE} & \multicolumn{1}{c}{ACIQ} & \multicolumn{1}{c|}{KL} & \BF{Best} & 0.01 & 0.02 & 0.05 & 0.01 & 0.02 & 0.05 \\
    \midrule
       \multicolumn{1}{c|}{\multirow{5}[0]{*}{\shortstack{VGG-16 BN\\(73.4)}}}
     & 8     & \BF{73.0} & 72.6  & 72.8  & 68.4  & 73.0  & 72.6  & 72.9  & 72.8  & 72.7  & 72.8  & 72.5 \\
     & 7     & \BF{72.8} & 72.5  & 72.5  & 60.7  & 72.8  & 72.1  & \BF{72.8} & 72.5  & 72.4  & 72.1  & 72.6 \\
     & 6     & 70.8  & \BF{71.3} & 71.2  & 63.2  & 71.3  & \HL{72.3} & 72.2  & 72.3  & \BF{71.8} & 71.8  & 72.1 \\
     & 5     & 63.1  & \BF{66.9} & 61.2  & 62.7  & 66.9  & \HL{69.3} & 70.2  & 71.0  & \HL{68.8} & 69.5  & 70.0 \\
     & 4     & 0.2   & 53.5  & 34.2  & \BF{59.4} & 59.4 & 10.4  & 26.3  & 37.9  & \HL{63.8}  & 63.8 &  65.9 \\
    \midrule
    \multicolumn{1}{c|}{\multirow{5}[0]{*}{\shortstack{ResNet-50\\(76.1)}}} & 8     & 75.4  & \BF{75.5} & 75.4  & 73.5  & 75.5  & \BF{75.7} & 75.7  & 75.7  & \BF{75.7} & 75.7  & 75.4 \\
     & 7     & 75.0  & \BF{75.2} & 75.0  & 72.8  & 75.2  & \BF{75.5} & 75.5  & 75.6  & \BF{75.5} & 75.5  & 75.5 \\
     & 6     & 72.9  & 73.5  & \BF{74.3} & 71.6  & 74.3  & \BF{74.9} & 74.7  & 75.0  & \BF{74.8} & 74.8  & 75.2 \\
     & 5     & 14.5  & 69.1  & \BF{69.9} & 69.4  & 69.9  & 69.4  & \BF{71.9} & \HL{72.6} & \HL{71.0} & 71.9  & 73.4 \\
     & 4     & 0.1   & 45.0  & 33.2  & \BF{62.9} & 62.9 & \multicolumn{1}{c}{12.1} & \multicolumn{1}{c}{36.1} & 55.2  & \HL{66.2} & 67.1  & 69.3 \\
 \midrule
 \multicolumn{1}{c|}{\multirow{5}[0]{*}{\shortstack{DenseNet-121\\(74.4)}}} & 8     & \BF{74.1} & 73.8  & 73.7  & 71.0 & 74.1  & \BF{74.2} & 74.2  & 74.2  & \BF{74.2} & 74.2  & 74.2 \\
      & 7     & \BF{73.8} & 73.3  & 73.1  & 62.3 & 73.8  & \BF{73.9} & 74.0  & 74.0  & \BF{74.1} & 74.2  & 74.1 \\
    & 6     & 71.0  & \BF{71.4} & 71.1  & 60.7 & 71.4  & \HL{72.9} & 73.0  & 73.2  & \HL{73.2} & 73.1  & 73.1 \\
    & 5     & 46.9  & \BF{65.4} & 61.4  & 54.6 & 65.4  & \BF{65.5} & \HL{69.7} & 71.3  & \HL{70.0} & 70.7  & 71.6 \\
     & 4     & 0.4   & 33.3  & 25.2  & \BF{42.6} & 42.6 & 13.1 & 37.5 & 53.0  & \HL{52.7} & 56.5  & 63.0 \\
 \midrule
 \multicolumn{1}{c|}{\multirow{5}[0]{*}{\shortstack{Inception-V3\\75.9)}}} & 8     & \BF{74.8} & 74.6  & 74.0  & 72.6  & 74.8  & \BF{75.2} & 75.4  & 75.3  & 74.8  & \BF{75.0} & 74.9 \\
    & 7     & \BF{73.2} & 71.2  & 69.1  & 69.4  & 73.2  & \HL{74.8} & 74.7  & 74.7  & 71.8  & \BF{73.8} & \HL{74.2} \\
    & 6     & 58.3  & \BF{66.2} & 62.3  & 63.0  & 66.2  & \HL{71.3} & 71.8  & 72.1  & \HL{70.5} & 71.7  & 72.5 \\
    & 5     & 0.5   & 30.4  & 29.6  & \BF{40.5} & 40.5  & \BF{45.2} & \HL{54.0} & 60.2  & \HL{57.0} & 60.0  & 62.9 \\
    & 4     & 0.1   & 0.2   & 0.1   & \BF{1.6} & 1.6   & 0.1   & 0.2   & 0.6   & \HL{2.1} & 2.3   & 4.8 \\
    \bottomrule
    \end{tabular}%
  \label{tab:ocs-weights}%
  \end{small}
  \vskip -0.1in
\end{table*}

This section reports experiments on CNN models for ImageNet classification~\citep{deng2009imagenet} conducted using PyTorch~\citep{paszke2017pytorch} and Intel's open-source Distiller~\footnote{https://github.com/NervanaSystems/distiller} quantization library.
Post-training quantization was performed using Distiller's symmetric linear quantizer, which scales the quantization grid based on the maximum absolute value following Equation~\ref{eqn:linearquant}.
For activation quantization, we first sampled the activation distributions using 512 \IT{training} images (i.e. images not part of the validation/test set) to determine the quantization grid points, then use this grid during testing. This profiling took between 40 and 200 seconds on our machine using an NVIDIA GTX 1080 Ti.
Weight clipping and OCS does not require profiling and was performed without any input data.

The chosen CNN benchmarks are four popular ImageNet classification models: VGG16~\citep{simonyan2015vgg} with batch normalization added, ResNet-50~\citep{he2015resnet}, DenseNet-121~\citep{huang2017densenet}, and Inception-V3~\citep{szegedy2015inception}.
Pre-trained weights were obtained from the PyTorch model zoo and we ran inference only.
The first layer was not quantized as it generally requires more bits than the others, and contains only 3 input channels meaning OCS would incur a large overhead.

\subsection{Effect of Quantization-Aware Splitting}
The first experiment compares our proposed quantization-aware (QA) splitting (Section~\ref{sec:q-aware}) against simply dividing by two as per Net2Net.
Table~\ref{tab:q-aware} displays results from ResNet-20 for CIFAR-10~\citep{krizhevsky2009cifar}.
Although the difference is negligible until 4 bits (at which point there is significant accuracy degradation), QA splitting is clearly better than the naive method. This validates our mathematical ideas in Section~\ref{sec:q-aware}, and we use QA splitting in all ensuing experiments.

% Table generated by Excel2LaTeX from sheet 'Paper'
\begin{table*}[tb]
\vskip -0.1in
\hfill
\begin{minipage}{0.7\linewidth}
  \centering
  \begin{small}
  \caption{\BF{ImageNet Top-1 validation accuracy with activation quantization --} formatting is identical to Table~\ref{tab:ocs-weights} except weight bits is kept at 8 while the activation bitwidth is changed. We did not combine OCS with clipping due to ineffectiveness of OCS on activations.
  }
  \vspace{3px}
    \begin{tabular}{c|c||cccc|c||ccc}
    \toprule
    \multicolumn{1}{c|}{\multirow{2}[0]{*}{\BF{Network}}} & \BF{Act.} & \multicolumn{4}{c|}{\BF{Clip}} & \BF{Clip} & \multicolumn{3}{c}{\BF{OCS}}  \\
    & \BF{Bits} & None & MSE & ACIQ & KL & \BF{Best} & 0.01 & 0.02 & 0.05 \\
    \midrule
    \multicolumn{1}{c|}{\multirow{5}[0]{*}{\shortstack{VGG16-BN\\(73.4)}}}
          & 8     & 72.5  & \BF{73.2} & 73.1  & \BF{73.2}  & 73.2 & 72.7  & 72.8  & 72.5 \\
          & 7     & 70.8  & \BF{72.8}  & \BF{72.8}  & 72.7 & 72.8 & 70.5  & 70.7  & 70.2 \\
          & 6     & 49.0  & 71.3  & \BF{71.4}  & 70.6      & 71.4 & 49.2  & 46.0  & 45.9 \\
          & 5     & 0.7  & \BF{62.0}  & 58.1  & 51.6       & 62.0 & 1.6  & 1.0  & 1.4 \\
          & 4     & 0.1  & \BF{11.5}  & 5.0   & 2.4        & 11.5 & 0.1  & 0.2  & 0.1 \\
    \midrule
    \multicolumn{1}{c|}{\multirow{5}[0]{*}{\shortstack{ResNet-50\\(76.1)}}} 
          & 8     & 75.5  & \BF{75.9} & 75.8  & 75.8        & 75.9 & 75.6  & 75.5  & 75.7 \\
          & 7     & \BF{75.4} & 75.3  & 75.2  & 75.3        & 75.4  & 74.1  & 74.5  & 74.1 \\
          & 6     & 62.6  & \BF{73.5} & \BF{73.5}   & 72.8  & 73.5  & 63.3  & 63.3  & 63.6 \\
          & 5     & 5.7   & 63.7 & \BF{65.4}   & 56.7       & 65.4  & 10.0  & 12.6  & 6.0 \\
          & 4     & 0.1   & 9.0 & \BF{20.6}   & 7.2         & 20.6 & 0.1   & 0.1   & 0.1 \\
    \midrule
    \multicolumn{1}{c|}{\multirow{5}[0]{*}{\shortstack{DenseNet-121\\(74.4)}}} 
          & 8     & 74.0  & \BF{74.1} & 73.8   & \BF{74.1} & 74.1             & 74.1   & \textit{\BF{74.2}} & 74.1 \\
          & 7     & 73.0  & \BF{73.7} & 72.9             & 73.7 & 73.6  & 73.2  & 73.2  & 73.0 \\
          & 6     & 67.2  & \BF{72.6} & 70.9  & 72.1     & 72.6 & 67.9  & 65.8  & 66.6 \\
          & 5     & 19.9  & \BF{66.9} & 64.6  & 64.5     & 66.9 & 16.0  & 18.7  & 13.2 \\
          & 4     & 0.2   & \BF{26.9} & 20.1  & 16.5     & 26.9 & 0.1   & 0.1   & 0.1 \\
    \midrule
    \multicolumn{1}{l|}{\multirow{5}[0]{*}{\shortstack{Inception-V3\\(75.9)}}}
          & 8     & 74.8  & \BF{75.1} & 73.4  & 75.0      & 75.1  & 74.8  & 74.9  & 74.8 \\
          & 7     & 72.6  & \BF{74.2} & 71.3   & 73.8     & 74.2  & 72.6  & 72.4  & 72.6 \\
          & 6     & 51.6  & \BF{69.6} & 60.7  & 67.9      & 69.6  & 54.1  & 51.5  & 48.5 \\
          & 5     & 1.3   & \BF{34.2} & 5.8  & 25.3       & 34.2  & 1.0   & 0.9   & 1.0 \\
          & 4     & 0.1   & \BF{0.3} & 0.1   & 0.2        & 0.3   & 0.1   & 0.1   & 0.2 \\
    \bottomrule
    \end{tabular}%
  \label{tab:ocs-acts}%
  \end{small}
\end{minipage}
\hfill
\begin{minipage}{0.29\linewidth}
  \centering
  \begin{small}
  \caption{\BF{ImageNet Top-1 accuracy for Oracle OCS on activations --} Oracle OCS splits a different set of channels for each input batch. Results use 6 activation bits and $r = 0.02$.
  }
  \vspace{3px}
    \begin{tabular}{c|cc}
        \toprule
        \BF{Batch}    & \BF{ResNet} & \BF{Inception} \\
        \BF{Size}     &  \BF{50}    & \BF{V3} \\
        \midrule
        1       & 74.6  & 71.7  \\
        2       & 74.5  & 71.7  \\
        4       & 74.0  & 71.6  \\
        8       & 74.1  & 70.9  \\
        32      & 73.5  & 70.7  \\
        128     & 73.3  & 70.3  \\
        \midrule
        No OCS & 62.6 & 51.6 \\
        \midrule
        Clip Best & 73.5 & 69.6 \\
        \bottomrule
    \end{tabular}
    \label{tab:ocs-acts-oracle}%
  \end{small}
\end{minipage}
\hfill
\vskip -0.1in
\end{table*}%

\subsection{Weight Quantization}
\label{sec:exp-weight}
Table~\ref{tab:ocs-weights} compares different clipping methods and OCS on weight quantization. 
The weights were quantized to 8-4 bits, while the activations were quantized to 8 bits.
Floating-point accuracy is displayed under the model name.
Linear quantization without clipping or OCS is shown in the \BF{Clip - None} column.
For ease of comparison we copy the best clipping result to the \BF{Clip - Best} column.
A range of small expand ratios $r$ was chosen for OCS.

Our results indicate that for large bitwidths, there is no advantage to doing weight clipping. This is in line with~\citep{tensorrt2017slides}, which reported the same at 8 bits.
%Indeed, on DenseNet-121 and Inception-V3 at 7 bits, clipping actually hurts accuracy performance despite improving on the stated metrics (MSE or KL divergence).
Clipping becomes beneficial at 6 bits or fewer, improving accuracy by up to $55\%$ for Resnet-50 and $40\%$ for Inception-V3.
Interestingly, the best-performing clipping technique depends on bitwidth and follows a consistent pattern across network architectures. As we go from high to low bitwidth, the winning technique goes from no clipping, to MSE/ACIQ, to KL at 4 bits.  %\fixme{add more here!}

Weight OCS with an expansion ratio of only $r=0.01$ outperforms our benchmarked clipping methods at 8-5 bits.
%Note that this is an unfair comparison --- we pit OCS again the best of the clipping methods, the latter being determined by peeking the test set.
At 8 and 7 bits the difference between OCS and clipping is small and there isn't a clear trend of improvement for higher expand ratios; in this regime OCS is not especially effective and the accuracy differences between expand ratios are mostly noise.
At 6 and 5 bits, OCS with $r=0.02$ outperforms clipping by $1\%$ for all models except ResNet-50, and up to $13\%$ for Inception-V3.
This demonstrates that the basic idea of OCS works --- by splitting the outliers to preserve their values instead of clipping them, OCS can improve the accuracy of post-training quantization.
Another trend is that OCS gets most of its gains from small expansion ratios. The gain from $r=0$ (no OCS) to $r=0.01$ is always larger than the gain from moving to higher $r$ values. Note that our benchmark networks have channel widths in the tens to hundreds so $r=0.01$ equates to a single channel split in many layers.
This again makes intuitive sense: the first channel split will target the unique largest outlier, guaranteeing a narrower weight distribution. Further splits target smaller values which occur with higher frequency, making OCS less effective at reducing the distribution width.

Given this intuition, we expect that a combination of OCS (to remove the largest outliers) followed by clipping (to further shrink the quantization grid) might surpass either method alone.
The rightmost columns of Table~\ref{tab:ocs-weights} show results for OCS + Best Clip (i.e. applying OCS followed by the best performing clip method at each bitwidth).
At 5 and 4 bits, OCS plus clipping cleanly outperforms OCS alone.
OCS and clipping both seek to shrink the dynamic range of the quantized values, and thus there is some level of overlap between them.
At high precision, OCS with our chosen expand ratios eliminates enough outliers such that additional clipping is unnecessary.
At low precision, we believe that OCS would require huge expand ratios to fully address the outlier problem --- in this regime OCS can combine with clipping to produce the best quantization results. 

%\fixme{MSE clips more aggressively than necessary}

\subsection{Activation Quantization}
\label{sec:exp-acts}
The same benchmarks and setup were used for activation quantization, except weights were kept at 8 bits while the bitwidth was varied for activations.
To select the channels to split, we sampled activation distributions and counted the number of extreme values (we used values greater than the 99'th percentile) in each channel. Channels with the highest counts were split.

Table~\ref{tab:ocs-acts} shows activation quantization results. Unlike the weights, clipping is effective at all bitwidths tested. This is again in agreement with~\citep{tensorrt2017slides}, which applied clipping to 8-bit activations.
MSE clipping outperforms the other clip threshold techniques in nearly all cases.
The gap between MSE and KL divergence is very small for large bitwidths, but at fewer bits MSE is clearly better.
ACIQ performs worse than the other two methods with the exception of ResNet-50, where it showed good performance. %This is expected as the technique was original proposed and evaluated for the weights only.

Activation OCS provides some improvement over simple linear quantization, but performs worse than clipping.
This is likely because OCS relies on being able to identify the exact channel containing the largest outlier. With activations, profiling can only indicate which channels are \IT{likely} to contain outliers, the best channel to split varies from input to input.
To test our explanation, we experiment with \BF{Oracle OCS}, which is simply OCS with exact knowledge of the activations generated by the network during testing --- Oracle OCS chooses different channels to split in each input batch.
Table~\ref{tab:ocs-acts-oracle} displays the results for Oracle OCS with different batch size on two models with 6 bit activations. Even at batch size 32, the oracle can already match or surpass the best clipping result. Further reducing the batch size (allowing channel selection at a finer granularity) leads to even better accuracy.
These results show that OCS and our channel selection strategy can be effective for activations.
However, channel selection must be done \IT{dynamically}, requiring additional run-time analysis which is difficult to implement and likely inefficient in commodity systems.

\begin{table}[tb]
  \centering
  %\vskip 0.05in
  \begin{small}
  \caption{\BF{Model size overhead for ResNet-50 with OCS} -- the overhead is very close to the user-provided expand ratio.}
    \begin{tabular}{c|cccc}
    \toprule
    \multirow{2}[0]{*}{\BF{ResNet-50}} & \multicolumn{4}{c}{\BF{OCS Expand Ratio}} \\
          & \BF{0.01} & \BF{0.02} & \BF{0.05} & \BF{0.1} \\
    \midrule
    Rel. Weight Size & 1.01  & 1.02  & 1.05  & 1.1 \\
    Rel. Activation Size & 1.02  & 1.03  & 1.06  & 1.11 \\
    \bottomrule
    \end{tabular}%
    \label{tab:overhead}%
  \end{small}
  \vskip -0.2in
\end{table}

\subsection{OCS Memory Overhead}
Because OCS increases the input channels by a factor of $r$, rounded up, the expand ratio $r$ is a lower bound for the model size overhead. Table~\ref{tab:overhead} shows both weight and activation overhead for ResNet-50 with different values of $r$, show that the true overhead matches $r$ very closely.

%In PyTorch, OCS is implemented by splitting the weights ahead of time and duplicating the activation channels during inference using \TT{torch.index\_select}.
%Figure~\ref{fig:runtime} plots the total wallclock inference time over the test set for various models, with weight OCS using various $r$ values. The baseline (i.e. $r=0$) is unmodified; only non-zero $r$ values run the additional OCS code.
%The time needed to preprocess the weights (1-2 minutes) is not included as what matters in practice is the latency of serving inference.
%The results show that OCS has negligible impact on inference latency in commodity hardware. Given the small memory overheads, we expect this to also hold true for specialized DNN accelerators.
\section{Experimental Evaluation on RNNs}

This section reports experiments on an RNN model with two stacked LSTM layers for language modeling~\citep{zaremba2014recurrent}. The corpus is the WikiText-2 dataset~\citep{merity2016pointer} with a vocabulary of 33,278 words. Each LSTM layer has a hidden size of 650, and the dimension of the word embedding in the input layer is 650.
As the CNN results have shown that activation OCS is not effective, we focused on OCS and clipping on the weights. Activations and the hidden state are kept in floating-point for this experiment.

\begin{table}[t]
  \vskip -0.05in
  \centering
  \begin{small}
  \caption{\BF{WikiText-2 perplexity with quantized weight --} lower is better. The floating-point baseline achieves a perplexity of 95.1. The best performing clip method along each row is bolded.}\vspace{3px}
    \begin{tabular}{c|c|cccc}
    \toprule
    \BF{Wt.} & \BF{Expand} & \multicolumn{4}{c}{\BF{Clip Method}} \\
    \BF{Bits} & \BF{Ratio} & None & MSE & ACIQ & KL \\
    \midrule
    \multicolumn{1}{c|}{\multirow{4}[0]{*}{6}}
          & 0.00 & \BF{94.5}  &	98.1  & 99.0	& 97.7 \\
          & 0.01 & \BF{95.0}  &	97.9  & 99.0	& 97.7 \\
          & 0.02 & \BF{94.6}  &	97.8  & 96.1	& 96.3 \\
          & 0.05 & \BF{93.9}  &	96.6  & 96.1	& 95.9 \\
    \midrule
    \multicolumn{1}{c|}{\multirow{4}[0]{*}{5}}
          & 0.00 & \BF{98.2}  & 99.4 & 100.8  & 98.8 \\
          & 0.01 & \BF{97.3}  & 99.7 & 99.9   & 97.7 \\
          & 0.02 & \BF{95.7}  & 98.9 & 99.2   & 97.0 \\
          & 0.05 & \BF{95.1}  & 98.2 & 98.5   & 96.3 \\
    \bottomrule
    \end{tabular}
  \end{small}
  \label{tab:ocs-rnn}
  \vskip -0.2in
\end{table}

Table~\ref{tab:ocs-rnn} compares the effects of OCS combined with different clipping methods on weight quantization.
Lower perplexity is better, and the baseline floating-point model achieves a perplexity of $95.1$.
The best result on each row (i.e. the best clipping method at each OCS expand ratio) is bolded.
Clipping is not effective on this model --- none of the clipping techniques achieve any perplexity improvement. 
OCS achieves a much better result. At 6 bits, OCS begins to outperform the baseline with $r=0.05$. At 5 bits, OCS sees steady perplexity decrease with successively larger expand ratios, clearly outperforming the best clipping result past $r=0.02$.
This is strong evidence that OCS can effectively improve post-training quantization beyond what can be achieved via clipping.
\section{Conclusions and Future Work}
\label{sec:conclusion}

We propose outlier channel splitting, a method to improve DNN quantization without retraining which can be applied on commodity hardware.
OCS splits channels in a layer to reduce the magnitude of outliers.
Unlike the existing clip-based methods, OCS introduces a new tradeoff by reducing quantization error at the cost of network size overhead.
Experimental results demonstrate that OCS on weights outperforms state-of-the-art clipping techniques with minimal overhead on deep CNN and RNN benchmarks.
At very low precision, OCS in conjunction with clipping outperforms either method alone.
Because clipping is used in NVIDIA TensorRT --- a commercial post-training quantization flow --- we believe that OCS has potential applicability in real-life systems.

Future work includes a more in-depth study into different channel selection methods, as well as applying OCS quantization during training.
Specifically, we believe that OCS can help shape weight distributions during training to obtain better results than training for quantization alone.
\subsection*{Acknowledgments}
This work was supported in part by the Semiconductor Research Corporation (SRC) and DARPA. One of the Titan Xp GPUs used for this research was donated by NVIDIA.


\bibliography{ms}

\begin{thebibliography}{34}
\providecommand{\natexlab}[1]{#1}
\providecommand{\url}[1]{\texttt{#1}}
\expandafter\ifx\csname urlstyle\endcsname\relax
  \providecommand{\doi}[1]{doi: #1}\else
  \providecommand{\doi}{doi: \begingroup \urlstyle{rm}\Url}\fi

\bibitem[Banner et~al.(2018)Banner, Nahshan, Hoffer, and
  Soudry]{banner2018aciq}
Banner, R., Nahshan, Y., Hoffer, E., and Soudry, D.
\newblock {ACIQ: Analytical Clipping for Integer Quantization of Neural
  Networks}.
\newblock \emph{arXiv e-print}, arXiv:1810.05723, Oct 2018.

\bibitem[Chen et~al.(2015)Chen, Li, Li, Lin, Wang, Wang, Xiao, Xu, Zhang, and
  Zhang]{chen2015mxnet}
Chen, T., Li, M., Li, Y., Lin, M., Wang, N., Wang, M., Xiao, T., Xu, B., Zhang,
  C., and Zhang, Z.
\newblock {MXNet: A Flexible and Efficient Machine Learning Library for
  Heterogeneous Distributed Systems}.
\newblock \emph{arXiv preprint}, arXiv:1512.01274, Dec 2015.

\bibitem[Chen et~al.(2016)Chen, Goodfellow, and Shlens]{chen2015net2net}
Chen, T., Goodfellow, I., and Shlens, J.
\newblock {Net2net: Accelerating Learning via Knowledge Transfer}.
\newblock \emph{Int'l Conf. on Learning Representations (ICLR)}, May 2016.

\bibitem[Choi et~al.(2018{\natexlab{a}})Choi, Chuang, Wang, Venkataramani,
  Srinivasan, and Gopalakrishnan]{choi2018qnn}
Choi, J., Chuang, P. I.-J., Wang, Z., Venkataramani, S., Srinivasan, V., and
  Gopalakrishnan, K.
\newblock {Bridging the Accuracy Gap for 2-bit Quantized Neural Networks
  (QNN)}.
\newblock \emph{arXiv e-print}, arXiv:1807.06964, Jul 2018{\natexlab{a}}.

\bibitem[Choi et~al.(2018{\natexlab{b}})Choi, Wang, Venkataramani, Chuang,
  Srinivasan, and Gopalakrishnan]{choi2018pact}
Choi, J., Wang, Z., Venkataramani, S., Chuang, P. I.-J., Srinivasan, V., and
  Gopalakrishnan, K.
\newblock {PACT: Parameterized Clipping Activation for Quantized Neural
  Networks}.
\newblock \emph{arXiv e-print}, arXiv:1805.0608, May 2018{\natexlab{b}}.

\bibitem[Chung et~al.(2018)Chung, Fowers, Ovtcharov, Papamichael, Caulfield,
  Massengill, Liu, Lo, Alkalay, Haselman, Abeydeera, Adams, Angepat, Boehn,
  Chiou, Firestein, Forin, Gatlin, Ghandi, Heil, Holohan, Husseini, Juhasz,
  Kagi, Kovvuri, Lanka, van Megen, Mukhortov, Patel, Perez, Rapsang, Reinhardt,
  Rouhani, Sapek, Seera, Shekar, Sridharan, Weisz, Woods, Xiao, Zhang, Zhao, ,
  and Burger]{microsoft2018brainwave}
Chung, E., Fowers, J., Ovtcharov, K., Papamichael, M., Caulfield, A.,
  Massengill, T., Liu, M., Lo, D., Alkalay, S., Haselman, M., Abeydeera, M.,
  Adams, L., Angepat, H., Boehn, C., Chiou, D., Firestein, O., Forin, A.,
  Gatlin, K.~S., Ghandi, M., Heil, S., Holohan, K., Husseini, A.~E., Juhasz,
  T., Kagi, K., Kovvuri, R.~K., Lanka, S., van Megen, F., Mukhortov, D., Patel,
  P., Perez, B., Rapsang, A.~G., Reinhardt, S.~K., Rouhani, B.~D., Sapek, A.,
  Seera, R., Shekar, S., Sridharan, B., Weisz, G., Woods, L., Xiao, P.~Y.,
  Zhang, D., Zhao, R., , and Burger, D.
\newblock {Serving DNNs in Real Time at Datacenter Scale with Project Brainwave
  }.
\newblock \emph{IEEE Micro}, 38\penalty0 (2):\penalty0 8--20, 2018.

\bibitem[Courbariaux et~al.(2015)Courbariaux, Bengio, and
  David]{courbariaux2015binaryconnect}
Courbariaux, M., Bengio, Y., and David, J.-P.
\newblock {BinaryConnect: Training Deep Neural Networks with binary weights
  during propagations}.
\newblock \emph{Advances in Neural Information Processing Systems (NIPS)}, pp.\
   3123--3131, 2015.

\bibitem[Deng et~al.(2009)Deng, Dong, Socher, Li, Li, and
  Fei-Fei]{deng2009imagenet}
Deng, J., Dong, W., Socher, R., Li, L.-J., Li, K., and Fei-Fei, L.
\newblock {ImageNet: A Large-Scale Hierarchical Image Database}.
\newblock \emph{Conf. on Computer Vision and Pattern Recognition (CVPR)}, pp.\
  248--255, 2009.

\bibitem[Han et~al.(2016)Han, Mao, and Dally]{han2016deep}
Han, S., Mao, H., and Dally, W.~J.
\newblock {Deep Compression: Compressing Deep Neural Networks with Pruning,
  Trained Quantization and Huffman Coding}.
\newblock \emph{Int'l Conf. on Learning Representations (ICLR)}, Feb 2016.

\bibitem[He et~al.(2015)He, Zhang, Ren, and Sun]{he2015resnet}
He, K., Zhang, X., Ren, S., and Sun, J.
\newblock {Deep Residual Learning for Image Recognition}.
\newblock \emph{arXiv e-print}, arXiv:1512.0338, Dec 2015.

\bibitem[Huang et~al.(2017)Huang, Liu, Weinberger, and van~der
  Maaten]{huang2017densenet}
Huang, G., Liu, Z., Weinberger, K.~Q., and van~der Maaten, L.
\newblock Densely connected convolutional networks.
\newblock \emph{Conf. on Computer Vision and Pattern Recognition (CVPR)},
  1\penalty0 (2):\penalty0 3, 2017.

\bibitem[Hubara et~al.(2017)Hubara, Courbariaux, Soudry, El-Yaniv, and
  Bengio]{hubara2017quantized}
Hubara, I., Courbariaux, M., Soudry, D., El-Yaniv, R., and Bengio, Y.
\newblock {Quantized Neural Networks: Training Neural Networks with Low
  Precision Weights and Activations}.
\newblock \emph{Journal of Machine Learning Research (JMLR)}, 18\penalty0
  (187):\penalty0 1--30, 2017.

\bibitem[Jacob et~al.(2018)Jacob, Kligys, Chen, Zhu, Tang, Howard, Adam, and
  Kalenichenko]{jacob2018quantization}
Jacob, B., Kligys, S., Chen, B., Zhu, M., Tang, M., Howard, A., Adam, H., and
  Kalenichenko, D.
\newblock {Quantization and Training of Neural Networks for Efficient
  Integer-Arithmetic-Only Inference}.
\newblock \emph{Conf. on Computer Vision and Pattern Recognition (CVPR)}, pp.\
  2704--2713, Jun 2018.

\bibitem[Jouppi et~al.(2017)Jouppi, Young, Patil, Patterson, Agrawal, Bajwa,
  Bates, Bhatia, Boden, Borchers, et~al.]{google2017tpu}
Jouppi, N.~P., Young, C., Patil, N., Patterson, D., Agrawal, G., Bajwa, R.,
  Bates, S., Bhatia, S., Boden, N., Borchers, A., et~al.
\newblock {In-Datacenter Performance Analysis of a Tensor Processing Unit}.
\newblock \emph{Int'l Symp. on Computer Architecture (ISCA)}, pp.\  1--12,
  2017.

\bibitem[Krizhevsky \& Hinton(2009)Krizhevsky and Hinton]{krizhevsky2009cifar}
Krizhevsky, A. and Hinton, G.
\newblock {Learning Multiple Layers of Features from Tiny Images}.
\newblock \emph{{Tech report}}, 2009.

\bibitem[Lin et~al.(2014)Lin, Maire, Belongie, Hays, Perona, Ramanan,
  Doll{\'a}r, and Zitnick]{lin2014coco}
Lin, T.-Y., Maire, M., Belongie, S., Hays, J., Perona, P., Ramanan, D.,
  Doll{\'a}r, P., and Zitnick, C.~L.
\newblock {Microsoft COCO: Common Objects in Context}.
\newblock \emph{European Conference on Computer Vision (ECCV)}, pp.\  740--755,
  2014.

\bibitem[McKinstry et~al.(2018)McKinstry, Esser, Appuswamy, Bablani, Arthur,
  Yildiz, and Modha]{mckinstry2018clip}
McKinstry, J.~L., Esser, S.~K., Appuswamy, R., Bablani, D., Arthur, J.~V.,
  Yildiz, I.~B., and Modha, D.~S.
\newblock Discovering low-precision networks close to full-precision networks
  for efficient embedded inference.
\newblock \emph{arXiv preprint}, arXiv:1809.04191, Sep 2018.

\bibitem[Merity et~al.(2016)Merity, Xiong, Bradbury, and
  Socher]{merity2016pointer}
Merity, S., Xiong, C., Bradbury, J., and Socher, R.
\newblock {Pointer Sentinel Mixture Models}.
\newblock \emph{arXiv preprint}, arXiv:1609.07843, Sep 2016.

\bibitem[Migacz(2017)]{tensorrt2017slides}
Migacz, S.
\newblock {8-bit Inference with TensorRT}.
\newblock \emph{NVIDIA GPU Technology Conference}, May 2017.

\bibitem[Park et~al.(2018{\natexlab{a}})Park, Kim, and Yoo]{park2018outlier}
Park, E., Kim, D., and Yoo, S.
\newblock {Energy-Efficient Neural Network Accelerator Based on Outlier-Aware
  Low-Precision Computation}.
\newblock \emph{Int'l Symp. on Computer Architecture (ISCA)}, Jun
  2018{\natexlab{a}}.

\bibitem[Park et~al.(2018{\natexlab{b}})Park, Yoo, and Vajda]{park2018value}
Park, E., Yoo, S., and Vajda, P.
\newblock {Value-aware Quantization for Training and Inference of Neural
  Networks}.
\newblock \emph{arXiv e-print}, arXiv:1804.07802, Apr 2018{\natexlab{b}}.

\bibitem[Park \& Choi(2019)Park and Choi]{park2019celldiv}
Park, H. and Choi, K.
\newblock {Cell Division: Weight Bit-Width Reduction Technique for
  Convolutional Neural Network Hardware Accelerators}.
\newblock \emph{Asia and South Pacific Design Automation Conf. (ASP-DAC)}, pp.\
   286--291, Jan 2019.

\bibitem[Paszke et~al.(2017)Paszke, Gross, Chintala, Chanan, Yang, DeVito, Lin,
  Desmaison, Antiga, and Lerer]{paszke2017pytorch}
Paszke, A., Gross, S., Chintala, S., Chanan, G., Yang, E., DeVito, Z., Lin, Z.,
  Desmaison, A., Antiga, L., and Lerer, A.
\newblock {Automatic Differentiation in PyTorch}.
\newblock \emph{Advances in Neural Information Processing Systems Workshops
  (NIPS-W)}, 2017.

\bibitem[Savchev \& Andreescu(2003)Savchev and Andreescu]{hermites}
Savchev, S. and Andreescu, T.
\newblock \emph{{Mathematical Miniatures}}, chapter 12. Hermite's Identity,
  pp.\  41--44.
\newblock Mathematical Association of America, 2003.

\bibitem[Settle et~al.(2018)Settle, Bollavaram, D'Alberto, Delaye, Fernandez,
  Fraser, Ng, Sirasao, and Wu]{settle2018quantizing}
Settle, S.~O., Bollavaram, M., D'Alberto, P., Delaye, E., Fernandez, O.,
  Fraser, N., Ng, A., Sirasao, A., and Wu, M.
\newblock {Quantizing Convolutional Neural Networks for Low-Power
  High-Throughput Inference Engines}.
\newblock \emph{arXiv preprint}, arXiv:1805.07941, May 2018.

\bibitem[Shin et~al.(2016)Shin, Hwang, and Sung]{shin2016fixed}
Shin, S., Hwang, K., and Sung, W.
\newblock {Fixed-Point Performance Analysis of Recurrent Neural Networks}.
\newblock \emph{Int'l Conf. on Acoustics, Speech and Signal Processing
  (ICASSP)}, pp.\  976--980, 2016.

\bibitem[Simonyan \& Zisserman(2015)Simonyan and Zisserman]{simonyan2015vgg}
Simonyan, K. and Zisserman, A.
\newblock {Very Deep Convolutional Networks for Large-Scale Image Recognition}.
\newblock \emph{arXiv e-print}, arXiv:1409.15568, Apr 2015.

\bibitem[Sung et~al.(2015)Sung, Shin, and Hwang]{sung2015resiliency}
Sung, W., Shin, S., and Hwang, K.
\newblock {Resiliency of Deep Neural Networks Under Quantization}.
\newblock \emph{arXiv preprint arXiv:1511.06488}, 2015.

\bibitem[Szegedy et~al.(2015)Szegedy, Liu, Jia, Sermanet, Reed, Anguelov,
  Erhan, Vanhoucke, and Rabinovich]{szegedy2015inception}
Szegedy, C., Liu, W., Jia, Y., Sermanet, P., Reed, S., Anguelov, D., Erhan, D.,
  Vanhoucke, V., and Rabinovich, A.
\newblock {Going Deeper with Convolutions}.
\newblock \emph{Conf. on Computer Vision and Pattern Recognition (CVPR)}, 2015.

\bibitem[Wu et~al.(2018)Wu, Li, Chen, and Shi]{wu2018training}
Wu, S., Li, G., Chen, F., and Shi, L.
\newblock {Training and Inference with Integers in Deep Neural Networks}.
\newblock \emph{Int'l Conf. on Learning Representations (ICLR)}, May 2018.

\bibitem[Xu et~al.(2018)Xu, Ding, Hu, Niemier, Cong, Hu, and
  Shi]{xu2018scaling}
Xu, X., Ding, Y., Hu, S.~X., Niemier, M., Cong, J., Hu, Y., and Shi, Y.
\newblock {Scaling for Edge Inference of Deep Neural Networks}.
\newblock \emph{Nature Electronics}, 1\penalty0 (4):\penalty0 216, 2018.

\bibitem[Zaremba et~al.(2014)Zaremba, Sutskever, and
  Vinyals]{zaremba2014recurrent}
Zaremba, W., Sutskever, I., and Vinyals, O.
\newblock Recurrent neural network regularization.
\newblock \emph{arXiv preprint arXiv:1409.2329}, 2014.

\bibitem[Zhou et~al.(2017)Zhou, Yao, Guo, Xu, and Chen]{zhou2017incremental}
Zhou, A., Yao, A., Guo, Y., Xu, L., and Chen, Y.
\newblock {Incremental Network Quantization: Towards Lossless CNNs with
  Low-Precision Weights}.
\newblock \emph{arXiv preprint}, arXiv:1702.03044, 2017.

\bibitem[Zhuang et~al.(2018)Zhuang, Shen, Tan, Liu, and
  Reid]{zhuang2018towards}
Zhuang, B., Shen, C., Tan, M., Liu, L., and Reid, I.
\newblock {Towards Effective Low-Bitwidth Convolutional Neural Networks}.
\newblock \emph{Conf. on Computer Vision and Pattern Recognition (CVPR)}, pp.\
  7920--7928, Jun 2018.

\end{thebibliography}
\bibliographystyle{icml2019}

\end{document}